\documentclass{Interspeech}

\interspeechcameraready

\title{Investigating the Impact of Word Informativeness on Speech Emotion Recognition}

\author[affiliation=]{Sofoklis}{Kakouros}

\affiliation[nocounter]{Department of Digital Humanities}{University of Helsinki}{Finland}
\email{sofoklis.kakouros@helsinki.fi}

\usepackage{comment}
\usepackage{multirow} 

\begin{document}

\maketitle

\begin{abstract}
    
    In emotion recognition from speech, a key challenge lies in identifying speech signal segments that carry the most relevant acoustic variations for discerning specific emotions. Traditional approaches compute functionals for features such as energy and F0 over entire sentences or longer speech portions, potentially missing essential fine-grained variation in the long-form statistics. This research investigates the use of word informativeness, derived from a pre-trained language model, to identify semantically important segments. Acoustic features are then computed exclusively for these identified segments, enhancing emotion recognition accuracy. The methodology utilizes standard acoustic prosodic features, their functionals, and self-supervised representations. Results indicate a notable improvement in recognition performance when features are computed on segments selected based on word informativeness, underscoring the effectiveness of this approach.
\end{abstract}

\section{Introduction}
Advancements in human-computer interaction (HCI) have significantly improved our interactions with technology, emphasizing the need for emotion-aware systems to facilitate more natural and effective communication, closer to human interactions. Speech emotion recognition (SER) has emerged as a key area within HCI, aiming to enable machines to detect and respond to users' emotional states through speech \cite{lee2005toward}. This pursuit is critical for developing interfaces that are more intuitive, suitable for a range of applications such as virtual assistants. Despite progress, accurately identifying emotions from speech remains challenging due to the subtle and complex nuances of emotional expression. These challenges range from theoretical, such as defining and understanding how emotions are elicited, expressed, and perceived \cite{brave2007emotion,picard2000affective}, to the practical aspects of detecting and interpreting these emotional cues in speech \cite{kakouros2023speech,stafylakis2023extracting}.


\subsection{Emotion recognition}
The challenges in SER are multifaceted but can generally be split into three main areas. First, the task demands the development of representations that can accurately and robustly encapsulate the acoustic variation indicative of various emotions. This has traditionally relied on engineered features like mel-frequency cepstral coefficients (MFCCs), filter banks (FBs) \cite{ ververidis2006emotional}, and sets of prosodic features such as the extended Geneva Minimalistic Acoustic Parameter Set (eGeMAPS) \cite{pepino2021emotion}. However, the emergence of self-supervised learning (SSL) methods such as HuBERT \cite{hsu2021hubert}, Wav2vec 2.0 \cite{baevski2020wav2vec}, and WavLM \cite{chen2021wavlm} has brought more refined representations that has led to state-of-the-art results in SER \cite{kakouros2023speech, yang2021superb,pepino2021emotion}. These innovative approaches represent a departure towards machine-driven learning of speech representations from unprocessed audio, offering a more intricate and layered representation of emotional expressions in speech.

The challenge of modeling the temporal dimension of emotions, which can manifest over short or extended speech sequences, constitutes the second major area. Emotions can be expressed across short segments such as a word but also extend to one or more utterances, requiring analytical strategies capable of recognizing and linking emotional cues over such prolonged periods. Techniques range from basic statistical pooling of time-based self-supervised representations \cite{yang2021superb} to advanced sequence modeling and the application of recurrent neural networks (RNNs), designed to map out and decode the emotional content of speech \cite{sarma2018emotion}. Further complicating SER efforts is the inherent ambiguity of emotional expressions, exacerbated by their multimodal aspects, including facial expressions and body language \cite{kim2015leveraging,mower2009interpreting}. This ambiguity often results in misinterpretation even among human judges, underscoring the necessity for SER methods that can disambiguate the indistinct and overlapping emotional cues. To address this, there are several approaches suggested in the literature such as specialized loss functions \cite{liu2021speech} and the modification of target labels using label smoothing \cite{kakouros2023speech}. This work focuses on addressing the second challenge (temporal chunking) to detect the segments in speech that carry the most important variation with respect to emotions.

\subsection{Predictive processing, LLMs, and SER}
Predictive processing offers a new lens through which to address the challenge of modeling the temporal dimension of emotions, emphasizing the role of linguistic expectations in interpreting acoustic information \cite{knill2004bayesian,rao1999predictive,kakouros2018making}. In general, according to this framework, the brain constantly generates predictions about incoming stimuli based on past experiences, adjusting these predictions as new information is received \cite{knill2004bayesian}. Large language models (LLMs), with their advanced understanding of linguistic structures and contexts, can simulate a similar predictive process for language. By analyzing the probability distributions of word sequences, LLMs can identify which parts of an utterance are predictable and which are surprising or informative. This capability to quantify surprisal in language opens new avenues for identifying salient segments of speech that may carry significant emotional weight.

Combining the insights from predictive processing with the analytical power of LLMs presents a novel approach to emotion recognition from speech. Instead of analyzing an entire utterance uniformly, this method proposes focusing on segments identified by LLMs as carrying high levels of surprisal—segments that deviate from the predictive linguistic model. These segments are likely to be more informative and carry acoustically more relevant variation to distinguish between emotions, as they reflect moments when the speaker has linguistically encoded and acoustically conveyed more information for the listener. By targeting these segments for acoustic analysis, the approach aims to capture the most relevant prosodic features and SSL representations that correlate with emotional expressions. This selective analysis could lead to more accurate and efficient emotion recognition, as it concentrates on parts of speech that are most indicative of the speaker's emotional state. The dataset for all experiments is publicly available and described in the next section. The code for all experiments in this paper is publicly available at \url{https://github.com/skakouros/predictive-emotion/}.

\section{The corpus and data processing}
In the next, the data used in the experiments are described as well as the feature extraction and post-processing of the features.
\subsection{RAVDESS}
The Ryerson Audio-Visual Database of Emotional Speech and Song (RAVDESS) is a multimodal database consisting of recordings from 24 professional actors (12 female) in North American accent \cite{ livingstone2018ryerson}. All actors enact the statements “Kids are talking by the door” and “Dogs are sitting by the door” with different emotions, at two intensity levels (normal and strong), and in speech and song conditions. Speech recordings were expressed with the emotions of calm, happy, sad, angry, fearful, surprise, disgust, and neutral. Song recordings were elicited with the emotions of calm, happy, sad, angry, fearful, and neutral. The dataset contains a total of 7356 recordings. Each recording was rated 10 times on emotional validity, intensity, and genuineness by 247 individuals. In the experiments only the speech portion of the corpus was used that contains 1440 audio files recorded by 12 male and 12 female actors. Following \cite{pepino2021emotion, venkataramanan2019emotion} the neutral and calm classes were merged, resulting into a total of 7 emotions.


\subsection{Feature extraction}
A standard set of acoustic features using eGeMAPS \cite{eyben2015geneva} was extracted. The feature set includes 88 different features and feature functionals including means and standard deviations of F0, loudness, spectral tilt, and MFCCs. The 88 eGeMAPS features and functionals are computed for each selected speech segment, with one single feature vector representing the entire segment.

To further evaluate potential separability of the emotion classes in terms of state-of-the-art self-supervised representations from speech, SSL representations were also extracted. SSL has been driving state-of-the-art results in many areas of speech technology, including areas where prosody is in question. Representations from SSL models are many orders of magnitude higher in dimensionality than standard acoustic features such as MFCCs, thus, their representation space, and their potential to capture relevant features in speech is much higher. SSL representations are extracted using \texttt{Wav2vec 2.0 base} where the utilized model checkpoint was hosted in the \texttt{Hugging Face} model library. The features were extracted from the pre-trained Wav2vec 2.0 by taking the representation from the last transformer layer of the model. To obtain an utterance-level description, the feature-level representations are pooled by taking the mean and standard deviation (std) of the features over each speech segment. This results into a $1536$ dimensional vector ($768$ mean and $768$ std concatenated) for each utterance in our data.


\section{Proposed method}
In the following section, the proposed methodology is described, focusing on the process of word selection. This includes an analysis of word surprisal, word rank, segment chunking, and the setup for supervised classification.

\subsection{Word surprisal}

Word surprisal is computed using predictions from GPT-2 small (the model is hosted in the \texttt{Hugging Face} model library)---GPT-2 models are trained on 40Gb of texts (WebText dataset). In particular, in the current experiments the GPT-2 small is used (\texttt{gpt2}; 124M parameters). For unigram computation publicly available word counts were used derived from the \texttt{Google Web Trillion Word Corpus}\footnote{https://norvig.com/ngrams/count\_1w.txt}.

The GPT family of models process a given text sequence using tokens. Tokens are common sequences of characters found in text. This means that a single word will not necessarily find a match in the model's dictionary and it might be split into a sequence of tokens. Tokenization is performed in order to reduce the model's dictionary size and to enable handling of out-of-vocabulary (OOV) words. The GPT models learn the statistical relationships between these tokens, and produce the next token in a sequence of tokens. Word surprisal can be seen as an information-theoretic measure of the amount of new information conveyed by a word \cite{hale2001,goodkind2018predictive}. In the current experiments, word surprisal values are extracted by taking the aggregate token surprisal over each word as follows:
\begin{equation}
\label{eq:total_surprisal}
{\bf w}_t = \sum_{\tau=0}^{N}{\bf S}_{\tau,L},
\end{equation}
\begin{equation}
\label{eq:word_surprisal}
{\bf S}_{\tau,L} = -log_2{\bf P}(w_{\tau}|w_{\tau-1,...,\tau-L}),
\end{equation}
where ${\bf S}_{\tau,L}$ is the surprisal value for token $\tau$ given $L$ previous tokens of word $t$ (${\bf w}_t$) that consists of $N$ tokens \cite{kakouros2023investigating,kakouros2016analyzing}.

\subsection{Word rank}
The \textit{rank} of a word refers to its position in a sorted list of all possible words, ordered by their predicted probabilities (from highest to lowest) by the LLM. In this work, the normalized rank is used instead. The normalized rank is an adjusted rank value to a scale that accounts for the size of the vocabulary of the LLM. This involves dividing the rank by the total number of words in the vocabulary, resulting in a value between 0 and 1, where a lower value indicates a higher probability (i.e., a higher ranking). Although word rank and probability are connected (probability indicates how likely a word is to be the next choice in a sequence, with a value between 0 and 1, whereas the rank is a relative measure that orders words by their probability) there is also a key difference: a word's rank can change based on the context and the probabilities of other words, even if its own probability remains the same.

\subsection{Segment selection}
In the experiments, speech segments of variable lengths were selected by sorting the words within each sentence according to their surprisal value or rank (e.g., from high to low surprisal). Specifically, for each sentence, words were ordered by their surprisal or rank, and the first $n$ words were isolated from the speech signal for subsequent feature extraction. Two distinct approaches were considered: (i) \textit{top-n}, wherein the speech segments corresponding to the top $n$ ordered words are concatenated and then subjected to feature extraction, and (ii) \textit{independent-n}, wherein only the speech segment of the word at position $n$, based on its surprisal or rank, is extracted from the sentence for feature extraction. These isolated segments serve as the basis for feature extraction.


\subsection{Supervised classification}

The eGeMAPS and Wav2vec 2.0 features were used to train a feed-forward Deep Neural Network (DNN)  classifier with emotions as the target classes. The data were selected to leave two unseen speakers out in the test set while the remaining speakers were used for training and validation. The splits were designed in order to have an equal proportion of female and male speakers in both the train and test sets. Each split of the dataset had a different set of speakers in the test set.

A feed-forward network with four hidden layers was used for training. The overall network layout is the following $L = [256, 128, 64, 32]$, where $L$ is the network layer. The network was trained for 100 epochs with a batch size 200 and learning rate of $lr$=1$e$-4. During training, dropout regularization was applied to the network to prevent overfitting. Specifically, a dropout rate of $p$=0.1 was applied to all layers except the last hidden layer. To optimize the network a binary cross-entropy loss function was used and adam optimizer.

\begin{table}[tbp]
\footnotesize 
\centering
\caption{Mean accuracy and F1 scores for the test set runs for unigram surprisal (SR), LLM surprisal, and rank-based word selection for Wav2vec 2.0. $top$ denotes the top $n$ words together while $ind$ the word independently at position $n$. Numbers in bold denote the best results in each metric and task.}
\vspace{-2mm}
\label{tab:wav2vec2results}
\begin{tabular}{ *{7}{c}} 
\hline
\multirow{2}{2em}{\textit{top}} & \multicolumn{2}{c}{\textbf{unigram SR}} & \multicolumn{2}{c}{\textbf{LLM SR}} & \multicolumn{2}{c}{\textbf{rank}} \\
  & \textit{ACC} & \textit{F1} & \textit{ACC} & \textit{F1} & \textit{ACC} & \textit{F1} \\
\hline
 1 & 57.81 & 56.15 & 60.97 & 59.28 & 57.31 & 55.43\\
 2 & 58.06 & 56.68 & 60.64 & 59.23 & 59.06 & 57.67\\
 3 & 61.65 & 60.75 & 60.81 & 59.46 & 62.39 & 60.94\\
 4 & 60.41 & 59.39 & \textbf{63.23} & \textbf{61.90} & 61.49 & 60.78\\
 5 & \textbf{62.48} & \textbf{61.20} & 61.15 & 59.71 & 62.07 & 60.56\\
 6 & 61.82 & 60.65 & 61.73 & 60.40 & \textbf{62.49} & \textbf{61.25}\\
\hline
\multirow{1}{2em}{\textit{ind}} & \textit{ACC} & \textit{F1} & \textit{ACC} & \textit{F1} & \textit{ACC} & \textit{F1} \\
\hline
 1 & \textbf{57.81} & \textbf{56.15} & \textbf{60.97} & \textbf{59.29} & 57.31 & 55.44\\
 2 & 57.22 & 55.72 & 55.74 & 53.76 & 58.13 & 56.31\\
 3 & 56.80 & 55.02 & 59.48 & 57.69 & 59.22 & 57.18\\
 4 & 13.28 & 3.35 & 13.28 & 3.35 & 13.28 & 3.35\\
 5 & 57.32 & 55.42 & 13.28 & 3.35 & 13.28 & 3.35\\
 6 & 13.28 & 3.35 & 13.28 & 3.35 & \textbf{59.81} & \textbf{58.06}\\
 \hline
\end{tabular}
\vspace{+1mm}
\end{table}

\section{Experiments}

\begin{table}[tbp]
\footnotesize 
\centering
\caption{Mean accuracy and F1 scores for the test set runs for unigram surprisal (SR), LLM surprisal, and rank-based word selection for eGeMAPS. $top$ denotes the top $n$ words together while $ind$ the word independently at position $n$. Numbers in bold denote the best results in each metric and task.}
\vspace{-2mm}
\label{tab:eGemMAPSresults}
\begin{tabular}{ *{7}{c}} 
\hline
\multirow{2}{2em}{\textit{top}} & \multicolumn{2}{c}{\textbf{unigram SR}} & \multicolumn{2}{c}{\textbf{LLM SR}} & \multicolumn{2}{c}{\textbf{rank}} \\
  & \textit{ACC} & \textit{F1} & \textit{ACC} & \textit{F1} & \textit{ACC} & \textit{F1} \\
\hline
 1 & 43.35 & 41.48 & 44.25 & 42.20 & 41.34 & 39.78\\
 2 & 46.26 & 44.79 & 46.28 & 44.42 & 45.10 & 43.32\\
 3 & 52.21 & 50.67 & 51.95 & 50.55 & 50.62 & 49.98\\
 4 & \textbf{54.23} & \textbf{53.31} & 50.95 & 50.19 & 50.38 & \textbf{50.09}\\
 5 & 53.13 & 51.64 & 49.46 & 48.23 & 47.61 & 46.64\\
 6 & 52.80 & 51.41 & \textbf{51.95} & \textbf{51.08} & \textbf{50.70} & 49.28\\
\hline
\multirow{1}{2em}{\textit{ind}} & \textit{ACC} & \textit{F1} & \textit{ACC} & \textit{F1} & \textit{ACC} & \textit{F1} \\
\hline
 1 & 43.35 & 41.48 & 44.25 & 42.20 & 41.34 & 39.78\\
 2 & \textbf{43.69} & \textbf{42.54} & 13.28 & 3.35 & 40.10 & 38.19\\
 3 & 41.68 & 39.76 & \textbf{44.85} & \textbf{43.53} & 13.28 & 3.35\\
 4 & 13.28 & 3.35 & 13.28 & 3.35 & 13.28 & 3.35\\
 5 & 13.28 & 3.35 & 13.28 & 3.35 & 13.28 & 3.35\\
 6 & 13.28 & 3.35 & 13.28 & 3.35 & \textbf{52.29} & \textbf{50.56}\\
 \hline
\end{tabular}
\vspace{+1mm}
\end{table}

A 10-fold cross-validation setup was used where at each fold two distinct speakers from the dataset were held-out for testing. Each held-out set consists of two speakers that are not present in the train and validation sets. This approach leaves approximately 10\% of the data for testing. Mean and standard deviation across folds is computed and presented as the aggregated result. The SER approach is evaluated with Wav2vec 2.0 and eGeMAPS.

Experiments were conducted for the top-n and independent-n cases and each for $n={1,2,3,4,5,6}$ where $n$ is the n-th word in a sentence. The outcomes of these experiments are presented in terms of the mean unweighted accuracies and F1 scores over all independent runs of the 10 folds.

\section{Results and analysis}
In the next the results are presented separately for the \textit{top n} and for the \textit{independent n} experiments. In the case of the different representations included in the experiments (Wav2vec 2.0 and eGeMAPS), a baseline result has been computed. The baseline represents the case where the entire utterance has been utilized for feature extraction, and each feature frame contributes to the aggregate vector computed for the entire utterance (as opposed to short segments selected based on the information value derived from the utilized metrics; e.g. unigram SR). Following feature extraction, the feature vectors were then used in a classification task in the same setup as the other tested conditions (top $n$ and independent $n$).

\subsection{top $n$}
An overview of the results for Wav2vec 2.0 and eGeMAPS can be seen in Tables \ref{tab:wav2vec2results} and \ref{tab:eGemMAPSresults}. The results show that individual words carry already a substantial proportion of the prosodic variation relevant for the identification of the individual emotion categories. Wav2vec 2.0 outperforms eGeMAPS as was expected and has been also earlier reported in the literature. For example, \cite{pepino2021emotion} report 57\% accuracy using eGeMAPS and 65.4\% with Wav2vec 2.0 on RAVDESS. These results are very close to the best performance obtained in the current setup, although different splits were used for the data (splits will be published on \texttt{github}).

Perhaps the most interesting finding is that few words seem to contain majority of the variation relevant for emotions, whereas including additional words does not seem to substantially improve performance. For example, in the case of Wav2vec 2.0 and when using LLM SR as the word selection criterion, emotion recognition performance using a single word already stands at 60.97\% accuracy. Inclusion of additional words does not push performance much further, with the best overall performance reached with the inclusion of the top four words being 63.23\% accuracy. The same behavior is also observed for eGeMAPS, although with overall lower performance. 

Another important finding that is consistent across the two tested features is that the baseline performance is lower when compared to that of using select words. This means that there is a certain degree of redundancy or non-pertinent variation for emotion recognition that drops the model performance when considering the entire utterance. Note that the baseline performance for Wav2vec 2.0 and eGeMAPS, when using the entire utterance, is 59.81\% and 52.3\%, respectively. In contrast, the best performance using selected words reaches 63.23\% and 54.23\% respectively.

\begin{figure}[t]
  \centering
  \includegraphics[width=\linewidth]{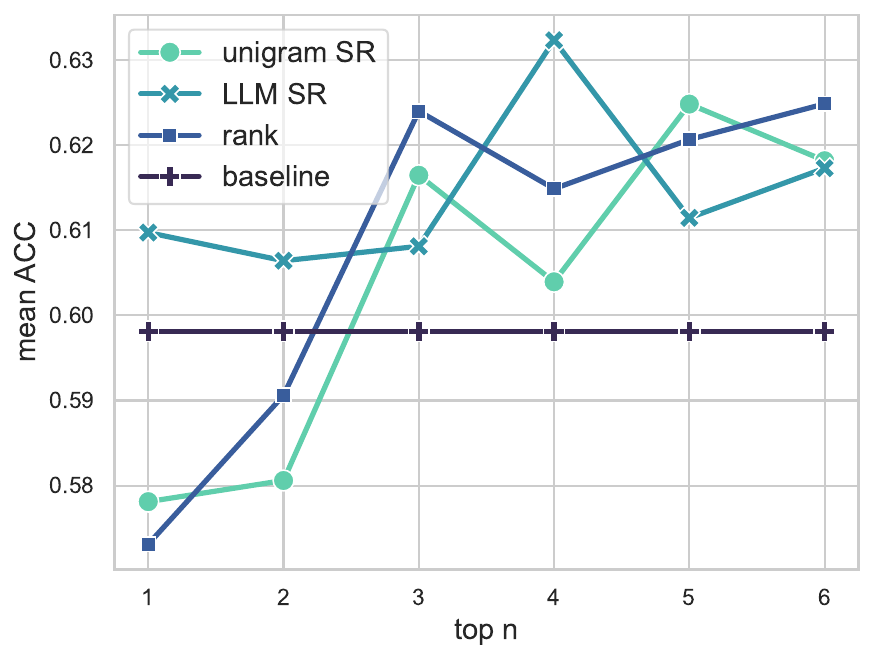}
  \caption{Mean accuracies across test runs for the top-$n$ experiments using Wav2Vec 2.0 representations. The horizontal reference line with mean $ACC = 0.59 (59.81\%)$ denotes the performance of the utterance-level baseline classifier.}
  \label{fig:meanvaluesWav2vec2}
\end{figure}

In terms of how the different word selection metrics perform, the results show that surprisal more effectively captures the prosodically and emotionally relevant information content for Wav2vec 2.0, giving the highest performance when isolated words are considered. In Figure \ref{fig:meanvaluesWav2vec2}, the results for Wav2vec 2.0 can be seen also with respect to the baseline performance. However, the result is not systematic across the tested features. When looking at eGeMAPS, the pattern is slightly different with no specific pattern observed to provide a better performance.

These findings raise the question of how do the ordered words perform independently? Also, what is the pertinent variation that gives this slight rise in performance? The first question is addressed in the next subsection.

\subsection{Independent $n$}

The results for independent $n$ can be seen in Tables \ref{tab:wav2vec2results} and \ref{tab:eGemMAPSresults}. Here too, the results show that individual words carry already a substantial proportion of the prosodic variation relevant for the identification of the individual emotion categories. For Wav2vec 2.0 and eGeMAPS, the words ordered closer to the top (e.g., first or second), meaning also that they are the most informative given their information content derived from the LLM, seem to provide also the best performance. In contrast, words appearing at the other end more often underperform leading to the model's lowest performance. For example, for Wav2vec 2.0, taking only the words that appear in position $n=1$ for LLM SR leads to 60.97\% accuracy whereas when the word at position $n=5$ or $n=6$ is taken leads to 13.28\% accuracy.

Among the different selection metrics used, the general observation is that unigram SR and LLM SR for both eGeMAPS and Wav2vec 2.0 provide useful insights into where the prosodically and emotionally relevant acoustic variation resides. This is evident from the systematically good performance that the models achieve in the top three ordered positions. In contrast, rank is not as informative with the performance being more widespread across the $n$ ordered positions.

The main assumption that was tested within this experiment is whether increased word informativeness would also be accompanied by increasing performance as reflected by the measured model's performance. In practical terms, the higher the $n$ the higher the accuracy would be expected to be. This was observed for Wav2vec 2.0 and partly for eGeMAPS where there was a similar pattern but not as systematic as in Wav2vec 2.0.

\section{Discussion}
In this work, a method was presented that leverages word informativeness, as determined by a pre-trained language model, to identify semantically significant segments in speech for emotion recognition. Traditional strategies, which calculate functionals for acoustic features like energy and F0 across entire sentences or larger speech portions, often overlook crucial fine-grained variations. By focusing on these semantically important segments, the proposed approach computes features solely within these areas, thereby enhancing the accuracy of emotion recognition. The methodology incorporates the use of standard acoustic prosodic features, their functionals, and self-supervised representations. The results demonstrate an improvement in recognition performance when features are computed on segments identified by word informativeness, highlighting the effectiveness of the method.

In particular, the results suggest that using word informativeness as extracted from an LLM such as GPT-2 can lead to improved performance. In addition, it was also found that using the few most informative words in a sentence to extract acoustic features is sufficient, leading to performance close to that of processing the entire utterance. This has important implications in the computational resources needed to process speech before training SER models. In addition, it was observed that using few words (utilizing a word selection criterion based on surprisal) rather than the entire utterance was more beneficial for model training leading to improved performance. In all, word informativeness seems to provide a useful tool in the aid of temporal segment selection from speech for SER.

However, this work also leaves some questions unanswered. For example, what is the additional information that the model leverages from including more words in the selection? One future avenue in this direction is to investigate what the eGeMAPS model learns with the inclusion of increasing words. Moreover, in this work, GPT-2 small was utilized. Perhaps larger and more complex models can provide more robust measures of word informativeness. Similarly, different and larger self-supervised learning speech models can be utilized to investigate potential differences in performance.

\section{Acknowledgements}

\ifinterspeechfinal
     This work was supported by the Research Council of Finland project no. 340125 “Computational Modeling of Prosody in Speech”. The author wish to acknowledge CSC – IT Center for Science, Finland, for providing the computational resources.
\else
     \textit{Acknowledgements will appear in the camera-ready version, not in the version submitted for review.}
\fi

\bibliographystyle{IEEEtran}
\bibliography{mybib}

\begin{thebibliography}{10}
\providecommand{\url}[1]{#1}
\csname url@samestyle\endcsname
\providecommand{\newblock}{\relax}
\providecommand{\bibinfo}[2]{#2}
\providecommand{\BIBentrySTDinterwordspacing}{\spaceskip=0pt\relax}
\providecommand{\BIBentryALTinterwordstretchfactor}{4}
\providecommand{\BIBentryALTinterwordspacing}{\spaceskip=\fontdimen2\font plus
\BIBentryALTinterwordstretchfactor\fontdimen3\font minus \fontdimen4\font\relax}
\providecommand{\BIBforeignlanguage}[2]{{%
\expandafter\ifx\csname l@#1\endcsname\relax
\typeout{** WARNING: IEEEtran.bst: No hyphenation pattern has been}%
\typeout{** loaded for the language `#1'. Using the pattern for}%
\typeout{** the default language instead.}%
\else
\language=\csname l@#1\endcsname
\fi
#2}}
\providecommand{\BIBdecl}{\relax}
\BIBdecl

\bibitem{lee2005toward}
C.~M. Lee and S.~S. Narayanan, ``Toward detecting emotions in spoken dialogs,'' \emph{IEEE transactions on speech and audio processing}, vol.~13, no.~2, pp. 293--303, 2005.

\bibitem{brave2007emotion}
S.~Brave and C.~Nass, ``Emotion in human-computer interaction,'' in \emph{The human-computer interaction handbook}.\hskip 1em plus 0.5em minus 0.4em\relax CRC Press, 2007, pp. 103--118.

\bibitem{picard2000affective}
R.~W. Picard, \emph{Affective computing}.\hskip 1em plus 0.5em minus 0.4em\relax MIT press, 2000.

\bibitem{kakouros2023speech}
S.~Kakouros, T.~Stafylakis, L.~Mo{\v{s}}ner, and L.~Burget, ``Speech-based emotion recognition with self-supervised models using attentive channel-wise correlations and label smoothing,'' in \emph{ICASSP 2023-2023 IEEE International Conference on Acoustics, Speech and Signal Processing (ICASSP)}.\hskip 1em plus 0.5em minus 0.4em\relax IEEE, 2023, pp. 1--5.

\bibitem{stafylakis2023extracting}
T.~Stafylakis, L.~Mo{\v{s}}ner, S.~Kakouros, O.~Plchot, L.~Burget, and J.~{\'C}ernock{\`y}, ``Extracting speaker and emotion information from self-supervised speech models via channel-wise correlations,'' in \emph{2022 IEEE Spoken Language Technology Workshop (SLT)}.\hskip 1em plus 0.5em minus 0.4em\relax IEEE, 2023, pp. 1136--1143.

\bibitem{ververidis2006emotional}
D.~Ververidis and C.~Kotropoulos, ``Emotional speech recognition: Resources, features, and methods,'' \emph{Speech communication}, vol.~48, no.~9, pp. 1162--1181, 2006.

\bibitem{pepino2021emotion}
L.~Pepino, P.~Riera, and L.~Ferrer, ``Emotion recognition from speech using wav2vec 2.0 embeddings,'' \emph{arXiv preprint arXiv:2104.03502}, 2021.

\bibitem{hsu2021hubert}
W.-N. Hsu, B.~Bolte \emph{et~al.}, ``{HuBERT: Self-supervised speech representation learning by masked prediction of hidden units},'' \emph{IEEE/ACM Transactions on Audio, Speech, and Language Processing}, vol.~29, pp. 3451--3460, 2021.

\bibitem{baevski2020wav2vec}
A.~Baevski, Y.~Zhou \emph{et~al.}, ``wav2vec 2.0: A framework for self-supervised learning of speech representations,'' \emph{Advances in Neural Information Processing Systems}, vol.~33, pp. 12\,449--12\,460, 2020.

\bibitem{chen2021wavlm}
S.~Chen, C.~Wang \emph{et~al.}, ``{WavLM: Large-scale self-supervised pre-training for full stack speech processing},'' \emph{arXiv preprint arXiv:2110.13900}, 2021.

\bibitem{yang2021superb}
S.-w. Yang, P.-H. Chi \emph{et~al.}, ``{SUPERB: Speech processing universal performance benchmark},'' in \emph{Proceedings of Interspeech}, 2021.

\bibitem{sarma2018emotion}
M.~Sarma, P.~Ghahremani \emph{et~al.}, ``Emotion identification from raw speech signals using dnns.'' in \emph{Interspeech}, 2018, pp. 3097--3101.

\bibitem{kim2015leveraging}
Y.~Kim and E.~M. Provost, ``Leveraging inter-rater agreement for audio-visual emotion recognition,'' in \emph{2015 International Conference on Affective Computing and Intelligent Interaction (ACII)}.\hskip 1em plus 0.5em minus 0.4em\relax IEEE, 2015, pp. 553--559.

\bibitem{mower2009interpreting}
E.~Mower, A.~Metallinou \emph{et~al.}, ``Interpreting ambiguous emotional expressions,'' in \emph{2009 3rd International Conference on Affective Computing and Intelligent Interaction and Workshops}.\hskip 1em plus 0.5em minus 0.4em\relax IEEE, 2009, pp. 1--8.

\bibitem{liu2021speech}
J.~Liu and H.~Wang, ``A speech emotion recognition framework for better discrimination of confusions,'' in \emph{Interspeech}, 2021, pp. 4483--4487.

\bibitem{knill2004bayesian}
D.~C. Knill and A.~Pouget, ``The bayesian brain: the role of uncertainty in neural coding and computation,'' \emph{TRENDS in Neurosciences}, vol.~27, no.~12, pp. 712--719, 2004.

\bibitem{rao1999predictive}
R.~P. Rao and D.~H. Ballard, ``Predictive coding in the visual cortex: a functional interpretation of some extra-classical receptive-field effects,'' \emph{Nature neuroscience}, vol.~2, no.~1, pp. 79--87, 1999.

\bibitem{kakouros2018making}
S.~Kakouros, N.~Salminen, and O.~R{\"a}s{\"a}nen, ``Making predictable unpredictable with style--behavioral and electrophysiological evidence for the critical role of prosodic expectations in the perception of prominence in speech,'' \emph{Neuropsychologia}, vol. 109, pp. 181--199, 2018.

\bibitem{livingstone2018ryerson}
S.~R. Livingstone and F.~A. Russo, ``The ryerson audio-visual database of emotional speech and song (ravdess): A dynamic, multimodal set of facial and vocal expressions in north american english,'' \emph{PloS one}, vol.~13, no.~5, p. e0196391, 2018.

\bibitem{venkataramanan2019emotion}
K.~Venkataramanan and H.~R. Rajamohan, ``Emotion recognition from speech,'' \emph{arXiv preprint arXiv:1912.10458}, 2019.

\bibitem{eyben2015geneva}
F.~Eyben, K.~R. Scherer, B.~W. Schuller, J.~Sundberg, E.~Andr{\'e}, C.~Busso, L.~Y. Devillers, J.~Epps, P.~Laukka, S.~S. Narayanan \emph{et~al.}, ``The geneva minimalistic acoustic parameter set (gemaps) for voice research and affective computing,'' \emph{IEEE transactions on affective computing}, vol.~7, no.~2, pp. 190--202, 2015.

\bibitem{hale2001}
J.~Hale, ``A probabilistic earley parser as a psycholinguistic model,'' in \emph{Proceedings of the second meeting of the North American Chapter of the Association for Computational Linguistics on Language technologies}.\hskip 1em plus 0.5em minus 0.4em\relax Association for Computational Linguistics, 2001, pp. 1--8.

\bibitem{goodkind2018predictive}
A.~Goodkind and K.~Bicknell, ``Predictive power of word surprisal for reading times is a linear function of language model quality,'' in \emph{Proceedings of the 8th workshop on cognitive modeling and computational linguistics (CMCL 2018)}, 2018, pp. 10--18.

\bibitem{kakouros2023investigating}
S.~Kakouros, J.~{\v{S}}imko, M.~Vainio, and A.~Suni, ``Investigating the utility of surprisal from large language models for speech synthesis prosody,'' in \emph{Proceedings of the 12th ISCA Speech Synthesis Workshop (SSW)}, Grenoble, France, 2023, pp. 127--133.

\bibitem{kakouros2016analyzing}
S.~Kakouros, J.~Pelemans, L.~Verwimp, P.~Wambacq, and O.~R{\"a}s{\"a}nen, ``Analyzing the contribution of top-down lexical and bottom-up acoustic cues in the detection of sentence prominence,'' \emph{Proceedings Interspeech 2016}, vol.~8, pp. 1074--1078, 2016.

\end{thebibliography}

\end{document}